\documentclass[review,3p]{elsarticle-arxiv}
\usepackage{bm}
\usepackage[table,xcdraw]{xcolor}
\usepackage[utf8]{inputenc}
\usepackage{url}  
\usepackage{hyperref}
\usepackage[T1]{fontenc}
\usepackage{color, colortbl, soul}
\soulregister\cite7
\soulregister\ref7
\soulregister\pageref7
\soulregister\url7
\soulregister\textit7
\definecolor{lightyellow}{cmyk}{0,0,0.50,0}
\sethlcolor{lightyellow}
\definecolor{yellow}{cmyk}{0,0,0.50,0}
\usepackage[cmex10]{amsmath}
\usepackage{balance}
\usepackage{makeidx}
\usepackage{amssymb}
\usepackage{latexsym}
\usepackage{graphicx}
\usepackage[tight,footnotesize]{subfigure}
\usepackage{comment}
\usepackage{booktabs}
\usepackage{multirow}
\usepackage{comment}
\usepackage{longtable}
\usepackage{tabularx}
\definecolor{Gray}{gray}{0.75}
\definecolor{LGray}{gray}{0.95}
\definecolor{lightyellow}{cmyk}{0,0,0.50,0}
\sethlcolor{lightyellow}
\definecolor{yellow}{cmyk}{0,0,0.50,0}
\definecolor{darkblue}{RGB}{18,10,143}
\newcommand{\modif}[1]{#1}

\newlength{\barwidth}
\newcommand{\cut}[1]{}

\newcommand{\tripadvisor}{\textsf{Tripadvisor}}
\newcommand{\booking}{\textsf{Booking}}

\newsavebox\verbtestsetuno
\newsavebox\verbtestsetdue

\begin{document}
\title{A study on text-score disagreement in online
  reviews}

\author[iit]{Michela Fazzolari\corref{cor1}\fnref{fn1}}
\ead{michela.fazzolari@iit.cnr.it}
\author[pad]{Vittoria Cozza}
\ead{vittoria.cozza@dei.unipd.it}
\author[iit]{Marinella Petrocchi}
\ead{marinella.petrocchi@iit.cnr.it}
\author[dtu,rm1]{Angelo Spognardi}
\ead{angsp@dtu.dk}
\address[iit]{Istituto di Informatica e Telematica -- CNR, via G. Moruzzi 1, 56124 Pisa, Italy}
\address[pad]{Department of Information Engineering, University of Padua, via Gradenigo 6/b, 35131, Padua, Italy}
\address[dtu]{DTU Compute, Technical University of Denmark, Richard Petersens Plads, 2800 Lyngby, Denmark}
\address[rm1]{Dipartimento di Informatica, Sapienza Universit\`a di Roma, via Salaria 113, Rome, Italy}
\cortext[cor1]{Corresponding author}
\fntext[fn1]{Phone number: +39 050 315 3483}

\begin{abstract}

\paragraph{Background/Introduction} In this paper, we focus on online reviews \modif{and employ artificial intelligence tools, taken from the cognitive computing field, to help understanding the relationships between the textual part of the review and the assigned numerical score.} We move from the intuitions that 1) a set of textual reviews expressing different sentiments may feature the same score (and vice-versa); and 2) detecting and analyzing the mismatches between the review content and the actual score may benefit both service providers and consumers, by highlighting specific factors of satisfaction (and dissatisfaction) in texts. 

\paragraph{Methods} To prove the intuitions, \modif{we adopt sentiment analysis techniques and we concentrate on hotel reviews, to find polarity mismatches therein.} In particular, we first train a text classifier with a set of annotated hotel reviews, taken from the \booking\ website. Then, we analyze a large dataset, with around 160k hotel reviews collected from \tripadvisor, with the aim of detecting a polarity mismatch, indicating if the textual content of the review is in line, or not, with the associated score. 

\paragraph{Results} \modif{Using well established artificial intelligence techniques and} analyzing in depth the reviews featuring a mismatch between the text polarity and the score, we find that -- on a scale of five stars -- those reviews ranked with middle scores include a mixture of positive and negative aspects. 

\paragraph{Conclusions} The approach proposed here, beside acting as a polarity detector, provides an effective selection of reviews -- on an initial very large dataset -- that may allow both consumers and providers to focus directly on the review subset featuring a text/score disagreement, \modif{which conveniently convey to the user a summary of positive and negative features of the review target}. 
\end{abstract}
  

\maketitle

\section{Introduction}
Social media offer a plethora of textual data posted by online users about the most disparate topics, e.g., liking or criticisms on politicians and celebrities, feedback on public events, comparisons among product brands, and suggestions for vacations. 
In this work, we consider online hotel reviews and we focus on those review systems that allow a reviewer
to post both a textual description of the hotel and a numeric
score, which quite directly summarizes the reviewer's overall satisfaction towards the facility. 
Actually, many e-commerce and e-advise websites, among which major
players, like Amazon, Walmart and Yelp, enable such dual
functionality.
 
The presence of both review text and score conveys to consumers a significant amount of information, which could be exploited in different ways. On the one hand, the score conveniently acts as a direct indicator, guiding the consumer to a faster choice, without getting lost into the details naturally contained in the review text. On the other hand,  the richness and variety of the information included in the text is supposed to improve the consumer awareness, supporting her cognitive process and, ultimately, leading to a satisfying purchase decision. Relying either on the text, the score, or both of them, the implicit belief
is that a correspondence exists between the polarity expressed by the textual data and the numerical value associated to such data. 
Instead, in this paper, we consider the presence of a possible misalignment between the review text and the review score. As noticed in~\cite{Mudambi14}, such a misalignment could lead to the increase of the consumer cognitive processing costs, to sub-optimal purchase decisions, and, ultimately, to neutralize the utility of the review site. Here, we approach the issue under a different perspective and with the aim of positively exploiting hidden information  that may exist, for those reviews featuring the disagreement between the text and the score.

Our intuition is that misalignment can naturally occur, since users' opinions are greatly subjective and it can be difficult and reductive to summarize a whole experience with a single value. 
For instance, less demanding people will probably turn a blind eye on the furniture of a hotel room, leading to a higher numerical score than that given by a hard to please client, but it is possible that both the reviews feature the same textual
description about such furniture. The same holds even in the evaluation of an objective characteristic of a service (e.g., the number of flights' delays of an airline): different users, such as businessmen and young travellers, may have a different perspective.

Following this intuition, we evaluate the disagreement between the text of a review and the associated score. For our investigations, we consider a 
large dataset consisting of around 160k hotel reviews collected from \tripadvisor. 
To evaluate if a {\it mismatch} exists between the text and the review score in the \tripadvisor\ dataset, we carry out a polarity detection task,  where texts are classified as positive or negative~\cite{Liu12,OpinionFinder2005,Wilson:2005, Wilson:2009}. The {\it polarity mismatch} attribute (i.e., the information about the correspondence -- or not -- between the review text polarity and the review score) is computed by constructing a reliable classification model that leverages on state-of-the-art techniques for sentiment analysis~\cite{weiss2004,hotho-etal-ldv-2005} and exploits a labelled dataset from the  \booking\ website.
\modif{Thus, we leverage techniques inherited from the field of cognitive computing area (such as sentiment analysis techniques), with the specific goals of identifying and analyzing mismatches between text and score in online review platforms. To the best of our knowledge, it is the first time that such techniques have been considered for that specific task.
}

\paragraph{Main findings} Findings are as follows. For the dataset under investigation,

\begin{itemize}
    \item at around 12\% of reviews with an actual score of 1 and 2 have been classified as positive by the classifier;
    \item at around 5\% of reviews with an actual score of 4 and 5 have been classified as negative by the classifier;
    \item among the mismatched reviews (i.e., reviews for which the detected polarity of the review text is opposed to the review score), the majority of the mismatches happens for reviews with an associated score of 2 and 4, \modif{rather than for reviews with the highest and lowest score; in addition, by analyzing in detail the reviews with a mismatch, we find out that their texts present a mixture of positive and negative content;}
    \item reviews for which a mismatch is not detected contain only negative and positive aspects, respectively.
\end{itemize} 

The proposed approach allows to slim down the set of reviews to take into account, when searching significant aspects of the products being reviewed. Indeed, the mismatch classification provides a selection of reviews in which positive and negative aspects of a product are mixed. Such a base represents a meaningful and compact piece of information, useful to both providers and consumers. By only relying on that focused reviews subset, the former will benefit by adjusting, e.g., their product lines and advertisement campaigns. The latter may concentrate only on such subset for addressing their needs and matching their expectations.
\modif{We think that our novel approach can be applied in other scenarios as well, where a text is associated to a value from a fixed scale, like surveys, peer-reviews of academic papers and student grades evaluations.  This could lead to the design and development of  a cognitive computing platform that can help and guide to the identification and, possibly, mitigation of any mismatch that could arise when users generate their contents. The platform could also be used by service administrators, as a pre-filter to highlight the most ambiguous or unsettled contents, to be considered for further analysis or alternative evaluations. Furthermore, our experimental results can also provide an additional approach in the context of Human-Computer-Interaction field, to shed light on how humans interact on and perceive online review platforms. The ultimate goal is the identification and understanding of the reasons behind such kind of physiological anomalies -- the mismatches -- that characterize user-generated contents.}

The remainder of this paper is as follows. \modif{Section~\ref{sec:RW} discusses related work in the area of online reviews platforms,  approached with cognitive computation and, specifically, polarity detection techniques. In Section~\ref{sec:basic}, we detail the datasets used in our study. The construction of the polarity classification model and the evaluation of its performance is described in Section~\ref{subsec:polarity}. In Section~\ref{sec:eval}, the classification model previously learned is applied to the \tripadvisor\ dataset, in order to evaluate the polarity mismatch of each review.} Then, we quantify the detected mismatches over the whole dataset, and we focus on specific kinds of mismatches, by showing and discussing real examples from the dataset. We also give further hints for some useful applications of the mismatch detection process.  Finally, Section~\ref{sec:conclusions} concludes the paper. 

\section{Related work}\label{sec:RW}
E-advice technology offers a form of ``electronic word-of-mouth'', with
new potential for gathering valid suggestions that guides the
consumer's choice. Since some years, extensive and nationally representative surveys
have been carried out, ``to evaluate the specific
aspects of ratings information that affect people attitudes toward
e-commerce''. It is the case, e.g., of work in~\cite{Flanagin14}, which
highlights how people, while taking into accounts the average of
ratings for a product, still do not take care of the number of reviews
leading to that average. \modif{The high impact of reviews on consumers is also testified by the fact that 
 a positive (or negative) review about a product can be as effective as a recommendation by a friend\footnote{see, e.g.,  \url{https://www.brightlocal.com/learn/local-consumer-review-survey/} (All URLs accessed on June 7, 2017)}. Further, positive comments convey a series of strong benefits, like, e.g., an improvement in search engines' ranking, a stronger perception of trust, and increased sales~\cite{ghose2011estimating,sparks2011theimpact,vermeulen2009theimpact}.}

In this work, we explore online reviews to understand if the text reflects the associated score, i.e., if there exists a polarity mismatch between text and score. \modif{A polarity mismatch can be detected by first applying polarity detection techniques to the text, whose outcome is the evaluation of the text content as expressing a positive (or negative) sentiment, and, then, to compare such positivity (or negativity) with the score associated to that text.}

Polarity detection techniques fall under the wide umbrella of sentiment analysis~\cite{Liu12,CambriaIS16}. Several approaches have been proposed in the literature for polarity detection.  A significant branch rely on lexicon-based features, due to the availability of lexical resources for sentiment analysis, such as, e.g., the lexicons SenticNet, SentiWordNet and a Twitter opinion lexicon, proposed in~\cite{cambria2016senticnet,cambriaBook,Esuli06sentiwordnet,BravoMarquez201665}, respectively. Usually, lexicon-based approaches involve the extraction of term polarities from the sentiment lexicons and the aggregation of the single polarities to predict the overall sentiment of a piece of text. 

Concerning subjectivity in texts, i.e., those expressions representing opinions and speculations, work in~\cite{OpinionFinder2005} is one of the first studies to perform subjectivity analysis, to identify subjective sentences and their features. 
In the specific field of polarity detection applied to product reviews, work in~\cite{Baccianella09} assigns a numerical score to a textual description exploiting the SentiWordNet lexicon: the task is especially useful when a reviews platform only allows to leave a text as a review, without an associated numerical score. A more recent work in~\cite{Fang2015} considers analogous topics. \modif{Work in~\cite{Pandarachalil2015} proposes an unsupervised approach that involves the extraction of terms and slangs polarities from three sentiment lexicons and the aggregation of such scores to predict the overall sentiment of a tweet}. 
In~\cite{Wilson:2005, Wilson:2009, Muhammad201692}, the authors consider the {\it contextual polarity} of a word, i.e., the polarity acquired by the word contextually to the sentence in which it appears.
\modif{For a survey of sentiment analysis algorithms and applications, the interested reader can refer to~\cite{Medhat20141093}.} For the specific scenario of polarity evaluation and sentiment analysis in specific social networks, the interested reader can refer to the series of work in~\cite{SemEval15,SemEval16}, inherent to Twitter.

Still regarding opinion mining in reviews,  efforts have been spent to investigate aspect extraction, i.e., the association between the expressed opinion and the opinion target~\cite{Poria201642}, the analysis of scarce-resource languages, like the Singaporean English~\cite{Lo2016236}, and future emotional behaviours of interactive reviewers~\cite{Bu201660}. \modif{Work in~\cite{cogncomp-shoppingbehavior} evaluates the differences in preferences between American and Chinese users.}
%
%
%
%
%
%
Work in~\cite{Kasper12} combines information extraction with sentiment analysis to identify a topic (e.g., ``wifi'') from a review segment, to recognize the dimension through which the topic is evaluated in the review (e.g., ``fast'', ``free'', ``poor'', etc.) and to evaluate how that topic got rated within that review segment. A similar approach is proposed in~\cite{brody2010anunsupervised}, where the authors present an unsupervised system to infer salient aspects in online reviews, together with their sentiment polarity.
\modif{A more recent method to help with the word polarity disambiguation has been proposed in~\cite{cogncomp-polaritydisambiguation}. In this work, the authors define the problem with a probabilistic model and adopt Bayesian models to calculate the polarity probability of a given word within a given context.}
Specific applications of polarity detection can be found in \cite{Vural2013} and \cite{valdivia2013sentiment}. The first contribution describes an unsupervised method for polarity detection in Turkish movie reviews, while the latter aims at detecting polarity of a Spanish corpus of movie reviews by combining a supervised and unsupervised learning in order to develop a polarity classification system.  \modif{Similarly, in~\cite{cogncomp-semanticparsing}, the authors rely on labeled reviews corpora  to test a novel approach for sentiment analysis, based on semantic relationships between words in natural language texts.}

\modif{This brief overview of the literature shows heterogeneous techniques and applications for polarity detection, both supervised and unsupervised, in different contexts and for different goals. In this work, thanks to the availability of a labeled dataset, we exploit a supervised approach, which automatically learns a model from the annotated data. In order to choose the most effective algorithm, we test different supervised algorithms and we finally select a linear Support Vector Machine (SVM)~\cite{Cortes1995}, due to its efficiency in dealing with the task at hand.}

\section{Datasets}\label{sec:basic}
\modif{In this section, we present the datasets used for our analysis. We consider two datasets, both composed of hotel reviews, downloaded from two popular e-advice sites, namely Booking\footnote{\url{http://www.booking.com}} and TripAdvisor\footnote{\url{http://www.tripadvisor.com}}.} The first, \modif{labeled, dataset is used to train a text classifier}, to learn the polarity of the reviews constituting it. The second one is not annotated and it is the input of the learned model. \modif{Both datasets were collected by developing ad-hoc software, which crawls the web pages of the hotels and extracts the reviews data.}

\subsection{\modif{Booking Labeled Dataset}}
\label{ref:dcap}
\modif{In order to train a text classifier,} we rely on a specific dataset, \modif{i.e., the Booking Labeled Dataset in Figure~\ref{fig:training}, which is focused on the hotel reviews domain}. \modif{To this end, we downloaded \modif{726,327} reviews and the associated metadata from the \booking\ website, by considering all the hotels reviews available for the city of London until July, 2016. Then, we filtered out all the reviews shorter than 20 words and written in a language different from English, by exploiting the \textit{language detection} Python library\footnote{\url{https://pypi.python.org/pypi/langdetect}}. We finally obtained 467,863 reviews.}
\modif{To tag each review with its \textit{strong} positive (or negative) polarity, we applied the following procedure:}
\begin{itemize}
    \item For each review, we considered the text content and its score. Since the \booking\ scoring system ranges over \{0, \ldots, 10\}, we discarded those reviews with a ``close-to-neutral" score, namely between 4 and 8.
    \item The remaining reviews were  tagged with a positive polarity if their score is above 8 and with a negative polarity if the score is below 4.
    \item We then manually inspect each review, to assess if the text content is in line with the \modif{polarity assigned in the previous step}.
    \item We finally keep 2,000 reviews for each polarity, to speed up the learning process of the classification model.
\end{itemize}

\modif{Thus, the Booking Labeled Dataset includes 4,000 reviews, half tagged with a strong positive polarity and the remaining half tagged with a strong negative polarity.} 

\subsection{\modif{TripAdvisor Dataset}}
\label{sec:ta_dataset}
The \tripadvisor\ dataset is composed of reviews taken from the
\tripadvisor\  website. This dataset contains all the reviews that could be accessed
on the website between the 26th of June 2013 and the 25th of June 2014 -- date of the newest extracted review -- for hotels in New York, Rome, Paris, Rio de Janeiro, and Tokyo. With a straightforward approach, we were able to collect the following
information, for each review:
\begin{itemize}
\item  the review date, text, and numeric score;
\item the reviewer username, location, and \textit{TripType}, being
  one among the following five categories: Family,
  Friends, Couple, Solo Traveler, and Businessman;
\item  the ID of the hotel which the review refers to.
\end{itemize}

We focus on the text and the score associated to each review. \modif{The reviews accessible from \tripadvisor\ in the year under investigation are 353,167. Nevertheless, we apply a filtering process to discard reviews whose textual part is not in English}, since the approach presented in Section~\ref{subsec:polarity} is specific for the English language. \modif{Reviews in English were selected by following the language identification and analysis approach presented in~\cite{Celli10}.}

\begin{figure}[hbt!]
 \centering
 \includegraphics[width=0.5\textwidth]{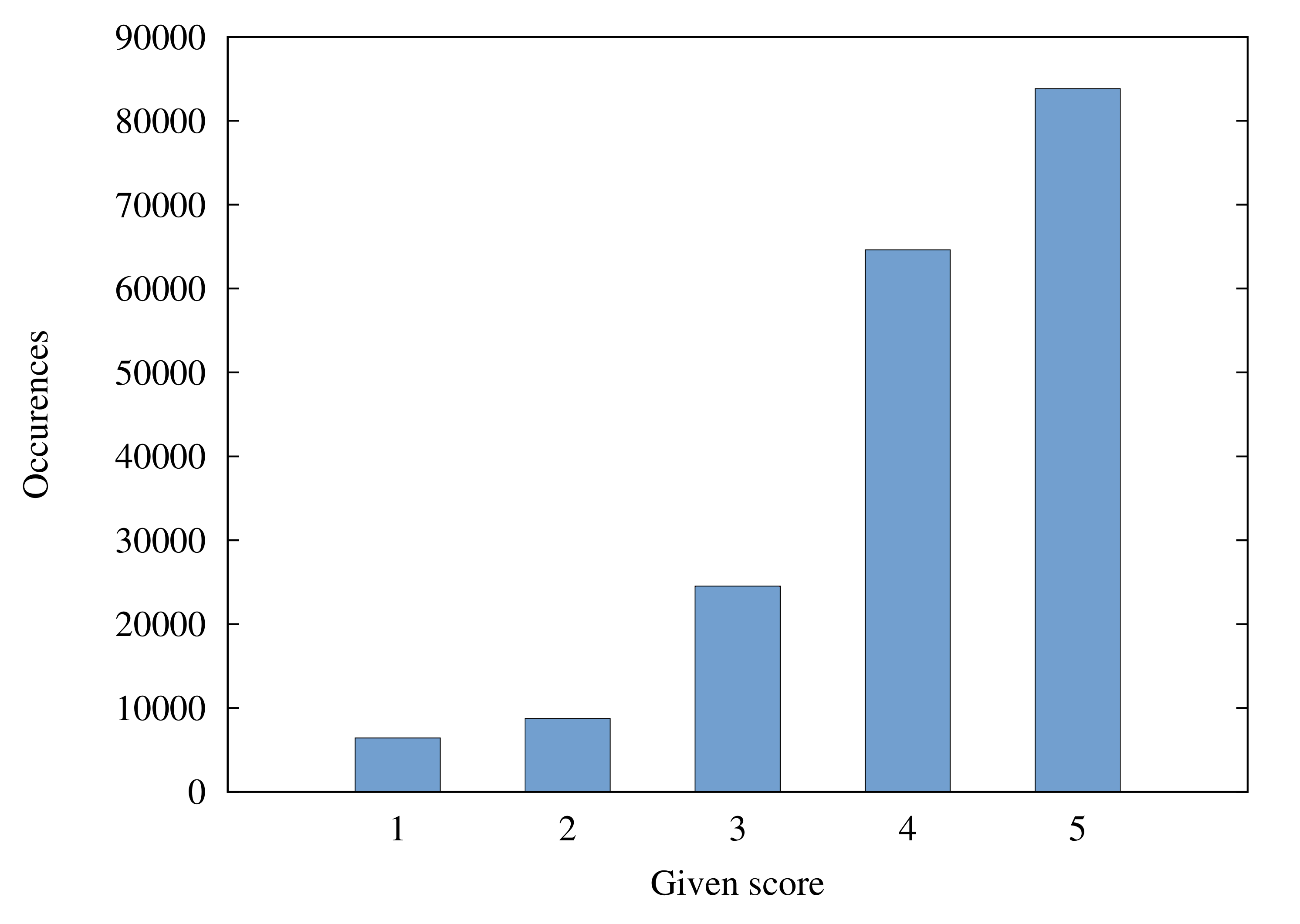}
 \caption{Distribution of scores in the \tripadvisor\ dataset.}
 \label{t1}
\end{figure}

Figure~\ref{t1} shows the
distribution of the reviews, per score value. Such distribution
 is highly unbalanced, with the highest score being the most frequent in the dataset (reflecting the distribution
usually featured by review platforms\footnote{Here an example of scores distribution for more than one million of reviews about electronic products on Amazon: \url{https://goo.gl/Es6L40}}).  Since in Section~\ref{subsec:polarity} we will focus on  \textit{strong disagreements}, we further discarded from the \tripadvisor\ dataset  those reviews having a score equal to 3.
\modif{Thus, after applying the filtering process -- by removing non English reviews -- and discarding the reviews with score equal to 3, the final dataset resulted made up of 164,300 reviews, in English, provided by 142,583 \tripadvisor's registered users that reviewed 4,019 hotels. }

\section{\modif{Polarity Classification Model Construction}}\label{subsec:polarity}
A {\it Polarity Mismatch} (PM) occurs when there is a disagreement between the text polarity of a review and the score assigned to it. In particular, here we focus on {\it strong disagreements}: on a scale of five stars, if a review text is evaluated as strongly negative, we expect the associated score to be 1 or 2 stars. Instead, if the text features a strongly positive polarity, we expect the score to be 4 or 5 stars. 

\modif{Given a set of reviews, our aim is to compute the PM for each of them, by performing a polarity recognition on the reviews' text.}  \modif{To this end, upon testing the performances of different classification algorithms, we adopt a linear Support Vector Machine (SVM)~\cite{Cortes1995} and we train the classifier on the Booking Labeled Dataset described in Section~\ref{ref:dcap}. We  use such dataset to learn a \textit{Polarity Classification Model} to automatically detect the polarity expressed by hotel reviews.}

\begin{figure}[ht]
\centering
\includegraphics[scale=0.4]{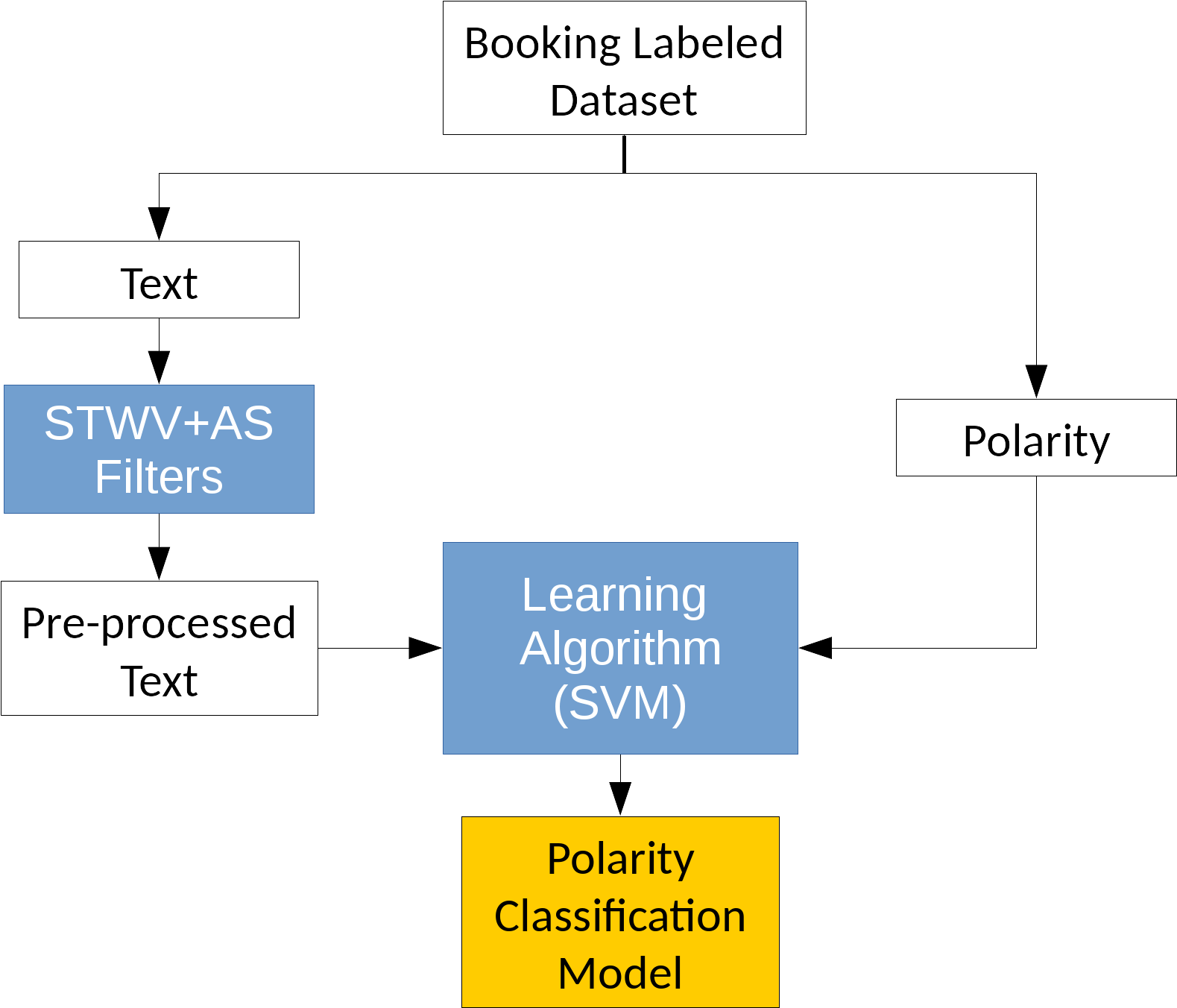}
\caption{Polarity Classification Model construction.}
\label{fig:training}
\end{figure}

\modif{The remainder of the section presents the steps performed to learn the Polarity Classification Model, how it has been tested and validated. These steps are summarized in Figure~\ref{fig:training}}.

\subsection{Text Filtering}\label{ref:txt_filtering}
Before building a classification model, the text needs to be pre-processed through Natural Language Processing (NLP) techniques \modif{and the most relevant features need to be selected. To this end, we exploit the String To Word Vector (STWV) and the Attribute Selection (AS) filters provided by Weka\footnote{\url{http://www.cs.waikato.ac.nz/ml/weka/}}. This step is represented in Figure~\ref{fig:training} by the building block \textit{STWV$+$AS Filters}, which returns a pre-processed review, given its text.}

The STWV filter supports all the common steps required in nearly every text classification task, like breaking text utterances into indexing terms (word stems, collocations) and assigning them a weight in term vectors. The STWV is an unsupervised filter that converts a text into a set of attributes representing the occurrences of the words in the text.

The set of words (attributes) is determined by the first batch filtered (typically training data). This filter has a significant number of parameters that can be set. In Table~\ref{tab:stwv_parameters} we report the parameters we have considered, with the chosen values, and their descriptions.

\begin{table}[ht!]
\centering
\caption{Parameter used for the STWV Weka filter.\label{tab:stwv_parameters}}
\begin{tabularx}{0.95\textwidth}{llX}
\toprule
\textbf{Parameter} & \textbf{Value} & \textbf{Description} \\
\midrule
TF Transform & True & The word frequencies are transformed into $f_{ij}*log$(num. documents/num. documents containing word i), where $f_{ij}$ is the frequency of word i in the j-th document. \\
\midrule
LowerCaseTokens & True & All the word tokens are converted to lower case before being added to the dictionary. \\
\midrule
OutputWordCounts & True & The filter outputs word counts rather than boolean 0 or 1 (indicating presence or absence of a word).\\
\midrule
Stemmer & Porter & We use the Porter stemmer as stemming algorithm, in order to reduce inflected (or sometimes derived) words to their word stem, base or root form.\\
\midrule
StopwordsHandler & WordsFromFile & We provided the filter with a text file containing a list of stopwords, which are words filtered out before natural language processing of data. \\
\midrule
Tokenizer & WordTokenizer & We use a simple word tokenizer and set the following delimiters: \textbackslash r\textbackslash n\textbackslash t.,;:'"()?!//W with //W referring to any special graphic character.\\
\midrule
WordsToKeep & 100,000,000 & This value allows the filter to keep all the possible useful attributes. \\
\bottomrule
\end{tabularx}
\end{table}

Keeping a large number of tokens as attributes, the STWV filter generates a huge number of attributes.  Therefore, we perform a dimensionality reduction, to transform the list of attributes into a more compact one and, also, to decrease the computational time required by the classification algorithms to build the models. To this end, we apply the Attribute Selection (AS) filter of Weka. This is a supervised filter, which selects a subset of the original representation attributes, according to some Information Theory quality metric. In Table~\ref{tab:as_parameters} we report the parameters used for the AS filter.

\begin{table}[ht!]
\centering
\caption{Parameter used for the AS Weka filter.\label{tab:as_parameters}}
\begin{tabularx}{0.95\textwidth}{llX}
\toprule
\textbf{Parameter} & \textbf{Value} & \textbf{Description} \\
\midrule
Evaluator & InfoGainAttributeEval & This is the metric used to evaluate the predictive properties of an attribute (or a set of them). We chose the Information Gain~\cite{Mitchell:1997:ML:541177}. \\
\midrule 
Search & Ranker & This is the search algorithm used to select the remaining group of attributes among all the available ones. We choose Ranker, which simply ranks the attributes according to the Information Gain metric and keeps those having the value above a predefined threshold. We set this threshold to 0, meaning that the filter keeps all those attributes scoring over 0. The Ranker evaluator has also a parameter that defines the number of attributes to keep. We left the default value (-1) meaning that all the useful attributed are selected.\\
\bottomrule
\end{tabularx}
\end{table}

\subsection{\modif{Classification Model Construction and Evaluation}}
\modif{The Booking Labeled Dataset is used to train several classification models, by exploiting different machine learning algorithms. In particular, we applied the  algorithm implementation of the Support Vector Machines (SVMs) described in~\cite{Platt1999}, the C4.5 decision tree algorithm~\cite{Quinlan1993}, the PART algorithm~\cite{Frank1998} and the Naive Bayes (NB) classifiers based on probabilistic classification algorithms~\cite{John1995}.}

The experiments were performed with the classifiers' parameters set to their default values \modif{in Weka} and a n-fold cross-validation methodology was applied, with n=5. \modif{In order to establish which classifier performs better, we evaluate and compare them by relying on} standard  metrics generally adopted for classification problems: Accuracy, Precision, Recall and F-score, whose computation is based on True Positive (TP), True Negative (TN), False Positive (FP) and False Negative (FN) values. Accuracy is the overall effectiveness of the classifier, i.e., the number of correctly classified patterns over the total number of patterns. Precision is the number of correctly classified patterns of a class over the number of patterns classified as belonging to that class. Recall is the number of correctly classified patterns of a class
over the number of samples of that class. The F-score 
is the weighted harmonic mean of precision and recall and it is used to compare different classifiers.

The average classification results are summarized in Table~\ref{tab:class_comparison}. Specifically, for each classifier, the table reports the accuracy and the per-class values of precision, recall and F-score. All the values are averaged over the 5 values obtained by applying the 5-fold cross validation. The best results have been obtained by SVM, with an average accuracy of 97.0\%. 

\begin{table}[htb!]
\centering
\caption{Comparison of classification results of different learning algorithms -- \booking\ dataset\label{tab:class_comparison}}
\begin{tabular}{lccccccc}
\toprule
\textbf{Classifier} & \textbf{Accuracy(\%)} & \multicolumn{2}{c}{\textbf{Precision(\%) by class}} & \multicolumn{2}{c}{\textbf{Recall(\%) by class}} & \multicolumn{2}{c}{\textbf{F-score(\%) by class}}\\
\cmidrule{3-8}
        &   & \textbf{pos} & \textbf{neg} & \textbf{pos} & \textbf{neg}& \textbf{pos} & \textbf{neg} \\
\midrule
SVM     & \textbf{97.00} & \textbf{96.9} & \textbf{97.1} & \textbf{97.1} & \textbf{96.9} & \textbf{97.0} & \textbf{97.0} \\
PART    & 92.25 & 91.5 & 93.1 & 93.2 & 91.3 & 92.3 & 92.2 \\
C4.5    & 90.15 & 90.0 & 90.3 & 90.4 & 90.0 & 90.2 & 90.1 \\
NB      & 89.95 & 94.9 & 86.0 & 84.4 & 95.5 & 89.4 & 90.5 \\
\bottomrule
\end{tabular}
\end{table}

\modif{Since the results obtained by the SVM classifier clearly outperforms those obtained by the other classifiers, we choose this algorithm (building block \textit{Learning Algorithm (SVM)} in Figure~\ref{fig:training}) to construct the Polarity Classification Model that will be used to detect the polarity of reviews belonging to the \tripadvisor\ dataset described in Section~\ref{sec:basic}. This  process is detailed in the following section.}

\section{\modif{Application of the Classification Model and Discussion}}\label{sec:eval}

\begin{figure}[ht]
\centering
\includegraphics[scale=0.4]{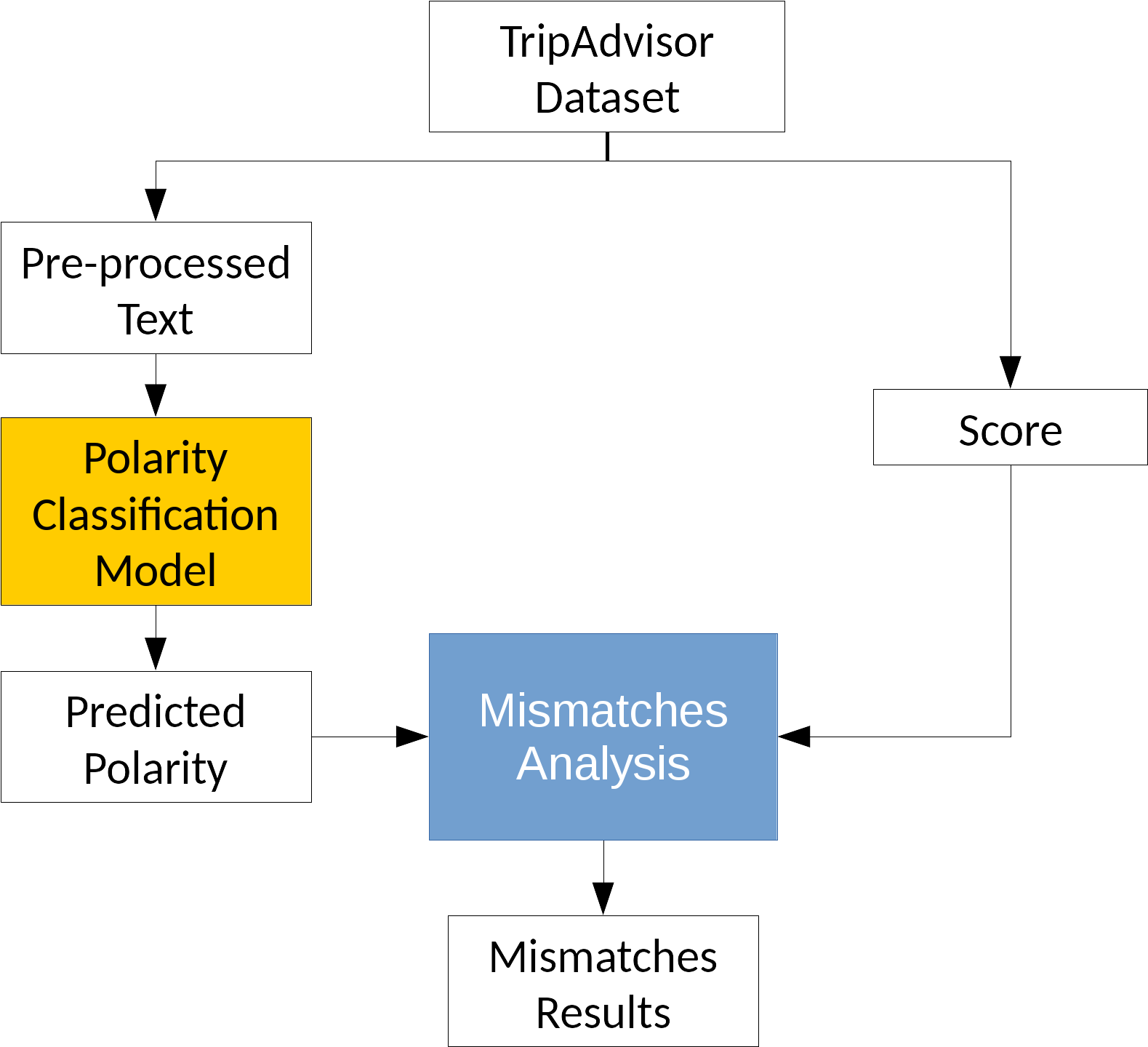}
\caption{\modif{Mismatches Detection approach.}
\label{fig:polDetect}}
\end{figure}

\modif{The Polarity Classification Model learned by using the Booking Labeled Dataset is here exploited to compute the PM for each review belonging to the \tripadvisor\ dataset. The followed approach is summarized in Figure~\ref{fig:polDetect}.}

\modif{The texts of the reviews are pre-processed, as already done for the \booking\ dataset, and then they are evaluated by the Polarity Classification Model, which outputs the predicted polarity for each review. The predicted polarity is finally compared with the actual polarity associated to the review text, which is extracted from the actual score assigned by the reviewer.}

Thus, if the detected polarity is positive and the review score is 4 or 5, then PM is set to 0 (no mismatch detected). Similarly, if the detected polarity is negative and the review score is 1 or 2, still PM is set to 0. Quite intuitively, if the detected polarity is positive, but the score is 1 or 2, then PM is set to 1. The same happens if the detected polarity is negative and the score is 4 or 5.

For the \tripadvisor\ dataset, the \modif{Polarity Classification Model} assigns PM equal to 0 to 94.07\% of patterns, i.e., in the 94.07\% of the cases, the predicted polarity matches the score associated to the review. Thus, it detects a PM for 9744 patterns, sum of False Positives (1806) and False Negatives (7938). Table~\ref{tab:conf_mat1} shows the confusion matrix for the SVM classifier.

\begin{table}[ht!]
\centering
\caption{Confusion matrix.\label{tab:conf_mat1}}
\begin{tabular}{ccrr}
\toprule
    & &\multicolumn{2}{c}{\textbf{Predicted Polarity}}\\
    & & pos & neg \\
\midrule
\textbf{Actual} & pos & 141097 & \bf 7938 \\
\textbf{Polarity}  & neg & \bf 1806 & 13459\\
\bottomrule
\end{tabular}
\end{table}

Table~\ref{tab:conf_mat2} expands the matrix by  grouping reviews according to their original score. The actual positive polarity spreads on 4 and 5 and the actual negative polarity spreads on 1 and 2.

\begin{table}[ht!]
\centering
\caption{Confusion matrix considering the actual score.\label{tab:conf_mat2}}
\begin{tabular}{ccrr}
\toprule
    & &\multicolumn{2}{c}{\textbf{Predicted Polarity}}\\
    & & pos & neg \\
\midrule
                & 5 & 81783	& \bf 2462 \\
\textbf{Actual} & 4 & 59314	& \bf 5476 \\
\cmidrule{2-4}
\textbf{Score}  & 2 & \bf 1522	& 7266 \\
                & 1 & \bf 284	& 6193 \\
\bottomrule
\end{tabular}
\end{table}

By looking at Table~\ref{tab:conf_mat2}, we can draw some considerations on the obtained results. Focusing on False Positives, the percentage of mismatches with score equal to 1 is 16\% and is considerably lower than the percentage of mismatches with score equal to 2. Similarly, if we consider the False Negatives, the percentage of mismatches with score equal to 5 is 31\%, which is lower than the percentage of mismatches with score equal to 4. Thus, the reviews with an intermediate score tend to be classified as a mismatch more than the reviews with extreme scores. 

Then, we compute the percentage of mismatched reviews with respect to each score and we report the results in Table~\ref{tab:mismatch_percentage}.
\begin{table}[ht!]
\centering
\caption{Mismatched reviews over the total number of reviews, per score.\label{tab:mismatch_percentage}}
\begin{tabular}{crr}
\toprule

\bf Score & \bf Total Reviews(\#) & \bf Mismatched Rev.(\%) \\
\midrule
 5 & 84245	& 3.0 \\
 4 & 64790	& 8.5 \\
 \midrule
 2 & 8788	& 17.3 \\
 1 & 6477	& 4.4 \\
\bottomrule
\end{tabular}
\end{table}

Table~\ref{tab:mismatch_percentage} highlights that the majority of mismatches happens when the score associated to the review is 2 and 4. Nevertheless, for the 2-score group, the percentage of mismatches is more than doubled, compared with the 4-score group. 

Considering the good performance of the SVM classifier on the annotated \booking\ dataset, reported in Table~\ref{tab:class_comparison}, we can reasonably assert that a relevant part of the mismatched reviews in the \tripadvisor\ dataset present, indeed, a PM between the text and the assigned score. To prove the outcome, we report some of the  mismatched reviews. Table~\ref{tab:false_positive} shows examples of False Positive reviews, scored with 1 and 2, but classified as positive by the classifier. For such an excerpt, we do not select ad-hoc reviews, rather we randomly select some examples from the available set.
It can be noticed that all the reviews include some positive words to describe positive aspects of the hotel (mainly the location, in the samples), so that the classifier is mistaken. 

\begin{footnotesize}
\begin{longtable}[htb!]{ccccp{9cm}}

\caption{False positive reviews (scores 1 and 2).}

\label{tab:false_positive}\\

\toprule
\textbf{\#} &
\textbf{Review} & 
\textbf{Actual} & 
\textbf{Predicted} & 
\textbf{Review}\\
  &
\textbf{Score} & 
\textbf{Polarity} & 
\textbf{Polarity} & 
\textbf{Text} \\
\midrule
\endfirsthead
\multicolumn{5}{r}{\footnotesize\itshape\tablename~\thetable: continues from the previous page} \\
\toprule
\textbf{\#} &
\textbf{Review} & 
\textbf{Actual} & 
\textbf{Predicted} & 
\textbf{Review}\\
  &
\textbf{Score} & 
\textbf{Polarity} & 
\textbf{Polarity} & 
\textbf{Text} \\
\midrule
\endhead
\midrule
\multicolumn{5}{r}{\footnotesize\itshape\tablename~\thetable: continues on the next page} \\
\endfoot
\bottomrule
\multicolumn{5}{r}{\footnotesize\itshape\tablename~\thetable: ends from the previous page} \\
\endlastfoot
1 & 1.0 & negative & positive & "We had a 5 night stay as a couple. This is a rundown tired beauty in a perfect location. The hotel has large rooms that have seen better days. The sheets were clean and the bed comfortable, however everything else needed cleaning and updating. It was advertised as air conditioned but it was not working!! Luckily we had French doors that allowed airflow. We did accept this room after being shown 3 others! The staff were rude and unhelpful. Wifi did work efficiently. The breakfast was reasonable. I would not stay again due to the staff who made life as difficult as possible for all guests."\\
\midrule
2 & 1.0 & negative & positive & "I stayed here and was bitten by something, simple as that. Not happy. I told the staff, they should do something about the bed, but they said nothing as if to deny there was a problem. Apart from the bites, the area is good, quiet and has some nice restaurants nearby. Also has good transport links, and a street market a short walk away."\\
\midrule
3 & 1.0 & negative & positive & "Incompetent front desk staff.  Overpriced breakfast. Teeny elevator that  fits one person and one bag. Rickety furniture in our room. Our seventh visit to Paris and this hotel was the worse.  Location is good and close St Germain de Pres."\\
\midrule
4 & 2.0 & negative & positive & "Very good location. Rooms relatively clean and comfortable. Beds not too bad. The tv was not working ok, so we had to switch rooms. Reception was helpful.  Not really three star, more like two.." \\
\midrule
5 & 2.0 & negative & positive & "The location of this hotel is good -- it is easy in and out to LaGuardia and JFK Airports via the Williamsburg Bridge. There is a City bike station (shared bicycle system) in the Roosevelt Park across the street. The room was spacious by New York standards. The room also has a full-sized kitchen which I did not use much because I was only staying for 2 nights. And the Wi-Fi in the room was unsteady." \\
\midrule
6 & 2.0 & negative & positive & "Staff tried to be useful but hardly was, hotel was dirty and dusty, and walls were thin as paper. However -- a good location (next to Place de la Republique), and a beautiful building (as all buildings in Paris)." \\
\bottomrule
\end{longtable}
\end{footnotesize}


In Table~\ref{tab:true_negative}, we report some examples of True Negative reviews, i.e., reviews scored with 1 and 2 and correctly detected as negative by the classifier. The excerpt highlights that these reviews essentially describe only negative aspects.
Also considering the numbers in Table~\ref{tab:conf_mat2}, we can argue that, among the reviews scored with 1 or 2, there exists a small subset featuring a PM. Such reviews mainly contain a mixture of positive and negative opinions, rather than only negative (or positive) ones.

\begin{footnotesize}
\begin{longtable}[htb!]{ccccp{9cm}}
\caption{True negative reviews (scores 1 and 2).}
\label{tab:true_negative}\\
\toprule
\textbf{\#} &
\textbf{Review} & 
\textbf{Actual} & 
\textbf{Predicted} & 
\textbf{Review}\\
  &
\textbf{Score} & 
\textbf{Polarity} & 
\textbf{Polarity} & 
\textbf{Text} \\
\midrule
\endfirsthead
\multicolumn{5}{r}{\footnotesize\itshape\tablename~\thetable: continues from the previous page} \\
\toprule
\textbf{\#} &
\textbf{Review} & 
\textbf{Actual} & 
\textbf{Predicted} & 
\textbf{Review}\\
  &
\textbf{Score} & 
\textbf{Polarity} & 
\textbf{Polarity} & 
\textbf{Text} \\
\midrule
\endhead
\midrule
\multicolumn{5}{r}{\footnotesize\itshape\tablename~\thetable: continues on the next page} \\
\endfoot
\bottomrule
\multicolumn{5}{r}{\footnotesize\itshape\tablename~\thetable: ends from the previous page} \\
\endlastfoot
1 & 1.0 & negative & negative & "They had to move us twice. Air conditioning sucks and rooms are dirty. They get ur cc just incase you use the condiments in the fridge but in ours their was stuff used and left over food in fridge which we told manager so she had truth that we was not going to be charged. I wish I had looked up on trip advisor before making my reservations." \\
\midrule
2 & 1.0 & negative & negative & "A very tired hotel, with more things broken in the room than the things that worked (bathroom extractor fan not working, broken furniture, door security chain missing etc). Our stay was topped by running a bath only to have brown matter flowing into the tub from the tabs. We of course reported this to the general manager, who was particularly disinterested,. We were directed to his office when we asked for him, after getting no response to the brown sludge problem from the rest of the staff, the general manager sat behind his desk and barely looked up, sweeping our concern aside by saying he would get someone to look at it. New York was fantastic but I would avoid at this hotel."\\
\midrule
3 & 1.0 & negative & negative & "Myself and my family stayed here at the end of October and I have to say I agree with the negative reviews below. The hotel rooms are tired the shower is nearly impossible to work and broken pipes means your shower to be flooded. I also found some of the reception staff to be particularly rude and arrogant towards us. The drinks and food are astonishingly high, especially as the bar/restaurant next door was fantastic and at least half the price. All in all a disappointment for us. We will be returning to New York but not to this hotel." \\
\midrule
4 & 2.0 & negative & negative & "Wrong choice to stay here!!!!! We travel in Rome on 20-23 Oct 13, we stay in this hotel, it was terrible to choose this hotel, I can\'t wait to check out from this hotel. Hotel is far from Roma attraction and subway station, they have shuttle bus but doesn't run often, sometime they cancel and leave hotel before time table. Around the hotel area is nothing, have small mall, no restaurant. The staff unfriendly, not helpful, rude service, never give information, you need to ask for all information and they don't speak english.They room was ok, breakfast just simple, food not so fresh.The wifi doesn't work well." \\
\midrule
5 & 2.0 & negative & negative & "Stay Somewhere Else. I recently stayed at the Quality Inn 08/08-08/11 and during my stay the ice machine was out of order and the wi-fi was not working for the first 2 days of my stay. It was a major inconvenience to walk down to another hotel just to use my laptop for work. I also made management aware of my issues and all I got back was I'm sorry so I will probably never use another choice hotel property in the future." \\
\midrule
6 & 2.0 & negative & negative & "Worst place I have ever stayed. Brutal. The shower had a 3 panel door with one missing. The bed felt like cardboard with springs pushing up. It was 2 twins pushed together with a gap between.  Breakfast was packed, no where to sit, and not very good. The toilet and sink are so close you can sit and brush your teeth at the same time! Tile floors throughout so every footstep sounded like a tap dancer.  Will never recommend this place and wont go back. They call themselves a 4 star hotel, this is a complete joke."\\

\bottomrule

\end{longtable}
\end{footnotesize}

We investigate  also the dual situation, by considering positive reviews, i.e. reviews marked with 4 or 5. To this end, Table~\ref{tab:false_negative} reports an extract of False Negative reviews, with a score equal to 4 or 5, but detected as negative by the classifier. From the table, we notice that these reviews globally express a positive sentiment, but customers highlight some issues within. 

\begin{footnotesize}
\begin{longtable}[htb!]{ccccp{9cm}}
\caption{False negative reviews (scores 4 and 5).}
\label{tab:false_negative}\\
\toprule
\textbf{\#} &
\textbf{Review} & 
\textbf{Actual} & 
\textbf{Predicted} & 
\textbf{Review}\\
  &
\textbf{Score} & 
\textbf{Polarity} & 
\textbf{Polarity} & 
\textbf{Text} \\
\midrule
\endfirsthead
\multicolumn{5}{r}{\footnotesize\itshape\tablename~\thetable: continues from the previous page} \\
\toprule
\textbf{\#} &
\textbf{Review} & 
\textbf{Actual} & 
\textbf{Predicted} & 
\textbf{Review}\\
  &
\textbf{Score} & 
\textbf{Polarity} & 
\textbf{Polarity} & 
\textbf{Text} \\
\midrule
\endhead
\midrule
\multicolumn{5}{r}{\footnotesize\itshape\tablename~\thetable: continues on the next page} \\
\endfoot
\bottomrule
\multicolumn{5}{r}{\footnotesize\itshape\tablename~\thetable: ends from the previous page} \\
\endlastfoot
1 & 4.0 & positive & negative & "The rooms are not as spacious as expected but it is definitely sufficient for a solo traveler. Location is great as you are literally just less than 100 meters away from the 35th St Herald Square subway station. Also it is strategically placed in Midtown for shoppers. Service was not too bad. I was only a little disappointed that upon my arrival I didn't have anyone to assist with my several pieces of luggage. The concierge was basically unmanned that time. Apart from that, I'm happy with the hotel overall for the price." \\
\midrule
2 & 4.0 & positive & negative & "Stayed for 2 nights recently on business. Was underwhelmed by the facilities. The lobby is very small, there's no restaurant (they use the restaurant next door), the elevators are tiny and the gym was an airless windowless room. If you arrive early they will not let you check in before 2, not so useful for a business traveler arriving from overseas and in need of changing facilities before going to a client meeting. We asked for  somewhere to change and were  directed to the gym -- where to my horror the only changing facility was a unisex bathroom, with no shower. The rates are competitive, but the rooms are typical sized for NYC. Very clean and compact, with in room Wifi."\\
\midrule
3 & 4.0 & positive & negative & "From the area I did not expect such a pretty hotel. Our room was small however we didn't plan on spending much time in it. The room had air conditioning, wifi, iPod Dock, plasma, slippers and robe. The only downside was the lighting, our room overlooked a small courtyard which didn't get much light.The hotel is 3 minutes from gare du nord station so easy reach of the metro, however we walked to many of the tourist areas such as Montmartre. We couldn't find any suitable restaurants near the hotel however we found a great bar called swinging londress around the corner." \\
\midrule
4 & 5.0 & positive & negative & "The rooms are small but adequate. The staff was great. The bed folds up during the day like a futon (there is a switch to make it go up and down). Being it is a futon the mattress was not real comfortable. I think next time we will choose a room with a real bed." \\
\midrule
5 & 5.0 & positive & negative & "Rooms on small side, but cozy and you can move around comfortably. They are well appointed, fast free Internet. Staff is accommodating, some hallways noise being heard nut not terrible at this point. Hotel is situated in a quieter part of herald sq yet close to the action. It's not the ritz, but I'd consider staying here instead if I was perfectly happy with just a little less space! Recommend"\\
\midrule
6 & 5.0 & positive & negative & "Tired! How to sleep in such a rumour???? It was horrible to hear all night bad rumours and noisy sounds from the installation and air conditioning! I couldn't sleep! Otherwise, it is a nice hotel with friendly staff!" \\
\bottomrule

\end{longtable}
\end{footnotesize}

Finally, we report some examples of True Positive reviews in Table~\ref{tab:true_positive}: such reviews are scored with 4 or 5 and they have been correctly classified as positives by the classifier. Overall, these reviews express a positive sentiment about the hotel they refer to, by describing only positive aspects.

\begin{footnotesize}
\begin{longtable}[htb!]{ccccp{9cm}}
\caption{True positive reviews (scores 4 and 5).}
\label{tab:true_positive}\\
\toprule
\textbf{\#} &
\textbf{Review} & 
\textbf{Actual} & 
\textbf{Predicted} & 
\textbf{Review}\\
  &
\textbf{Score} & 
\textbf{Polarity} & 
\textbf{Polarity} & 
\textbf{Text} \\
\midrule
\endfirsthead
\multicolumn{5}{r}{\footnotesize\itshape\tablename~\thetable: continues from the previous page} \\
\toprule
\textbf{\#} &
\textbf{Review} & 
\textbf{Actual} & 
\textbf{Predicted} & 
\textbf{Review}\\
  &
\textbf{Score} & 
\textbf{Polarity} & 
\textbf{Polarity} & 
\textbf{Text} \\
\midrule
\endhead
\midrule
\multicolumn{5}{r}{\footnotesize\itshape\tablename~\thetable: continues on the next page} \\
\endfoot
\bottomrule
\multicolumn{5}{r}{\footnotesize\itshape\tablename~\thetable: ends from the previous page} \\
\endlastfoot
1 & 4.0 & positive & positive &  "Great location. Few minutes walk to Broadway shows and times square. Close to tourist bus-stops, and also a subway stop. Convenient shop on site, and good little restaurant. Our room had all we needed. We had an amazing view of the Empire State Building out of our window.Staff were very friendly and helpful, and quick to help." \\
\midrule
2 & 4.0 & positive & positive & "I stayed here with two friends in June for one night. It is perfectly situated within walking distance of Gare du Nord and Gare de l'Est. Upon arrival, we were told that the elevator was out of order, so breakfast would be complementary, which I was very happy with. Our bags were carried up to our room, which was decorated beautifully.Breakfast in the morning was excellent, fresh bread and croissant with a selection of cheeses, cold meats and jams, as well as boiled eggs, cereal and a range of hot drinks. The staff stored our luggage for the day while we went exploring Paris at no extra charge." \\
\midrule
3 & 4.0 & positive & positive & "Very nice and well decorated, very conveniently located hotel, a street away from Notre Dame, RER/metro stations and Velib station. There are many restaurants, shops and bars on site. We would definitely stay in this hotel again."\\
\midrule
4 & 5.0 & positive & positive & "We stayed here over the valentines period and had an amazing time. Great staff Vanessa was extremely helpful in sharing information on the best places to visit in Paris. The room was extremely clean and very modern with a classic Parisian feel. Would definitely stay again and the breakfast was amazing!"\\
\midrule
5 & 5.0 & positive & positive & "Everything you would expect and more. From the warm greeting from Jonathan at the front desk to the wonderful view of the harbor when we entered the room. My wife and kids thoroughly enjoyed our two night visit. Our bellman, also Jonathan, was excellent and very knowledgeable about lower Manhattan. Much higher level of customer service than some of the larger hotels. Easy to catch a cab, a block and a half to the subway and right next to the best little cafe. We can't wait to come back!"\\
\midrule
6 & 5.0 & positive & positive & "I was lucky enough to stay here after a bit of bargain hunting online! A beautiful hotel, spacious rooms and a wonderful view -- an accessible balcony gave us a great view of the Chrysler building, along with the New York skyline!"\\
\bottomrule

\end{longtable}
\end{footnotesize}

Concluding, the classifier is able to detect reviews with a PM. The mismatch should not be intended as an inconsistency between the text written in the review and the score assigned to it; instead, it indicates that the considered review is a mixture of positive and negative opinions, to a greater extent if compared with reviews belonging to the same class of score. Thus, this approach features its benefits if exploited to perform an initial selection on reviews, by tagging with a mismatch those reviews which are worth to be further investigated in details.

\subsection{Open problems and further investigations} 

The approach presented so far let a specific subset of reviews emerge from a wider set. The characteristic of this subset is that the texts of the reviews contain a mixture of positive and negative aspects, leading the classification algorithm to label such reviews with a polarity in contrast to the associated score. 

This paves the way for further investigation. Indeed, even if in this work we only considered the relationships between the review text and the score associated to it, a further possibility could be to explore possible connections between the polarity mismatch and some characteristics of the reviewer (e.g., the gender) and the reviewed product (e.g., specific attractions in the hotel neighbourhood). As an example, the work in~\cite{Rahman15} studies how
review scores may be affected by external, environmental factors, such as the weather
conditions and the daylight length. Also, a series of recent studies correlates the huge amount of textual data available online with the demographic and psychological characteristics of the users who author them. This is the case, e.g., of the work proposed in~\cite{Schwartz:2013}, where the authors consider millions of Facebook messages, from where they extract hundreds of millions of words, topics, and sentences and automatically correlate them with gender, age, and personality of the users that posted them. Still referring to users' demographic characteristics, work in~\cite{Hovy2015} investigates textual online reviews, to test how the words -- and their use -- in a review are linked to the reviewer gender, country, and age. Work in~\cite{Minnich2015} focuses on review manipulation: comparing hotels reviews and related features
across different review sites, the work outperforms the detection of
suspicious hotel reviews when checking the reviews on sites in isolation.

Therefore, we envisage the possibility to extract additional features from the \tripadvisor\ dataset and to apply adequate techniques to discover frequent patterns, correlations and causal structures among them. To this aim, we may think to follow different approaches. One is represented by the well known and widely applied methodology of 
Association Rule Mining~\cite{Hipp00}, which allows the induction of rules
predicting the occurrence of one feature (or more), given the occurrence
of other features in the same set. This may lead to find correlations among the review features, such as the score, the occurrence of a possible polarity mismatch, and additional reviewers and reviews metadata. Furthermore, one could apply  statistical techniques of preferences measurement~\cite{Gustafsson2007}, largely applied in market analysis.  Among them, we remind the conjoint analysis~\cite{GreenRao1971,Netzer2008}, which aims at determining the combination of features that is mostly influential to choose a product. The goal, in our scenario, would be to recognize the most significant features leading to, e.g., a polarity mismatch. 

\section{Conclusions}\label{sec:conclusions}
In this work, we moved from the intuition that a misalignment can exist between a review text and the score associated to it. To prove this hypothesis, we first constructed a reliable classifier by using an annotated dataset of hotel reviews taken from \booking. We then used this classifier to classify a large dataset of around 160K hotel reviews taken from \tripadvisor, according to the positive or negative polarity expressed by their textual content. 

As the main result of our approach, we found that reviews tagged with a polarity mismatch present a mixture of positive and negative aspects of the product under examination. Thus, the mismatch classification is able to reduce the set of reviews which users may focus on, when searching significant aspects of the products being reviewed.

We argue that, only focusing on those texts associated with a mismatch, instead of manually investigating all the review texts in a dataset,  consumers could achieve a better awareness on what has been liked -- or not -- about a product. Also, providers could understand how to improve their services. The proposed technique is applicable to a wide range of services: accommodation, car rental, food services, just to cite a few. 

As future work, we first aim at running a semantic analysis on those texts of reviews marked as mismatches, focusing in particular on aspect extraction, to link the opinionated text with the target of the opinion. Secondly, we will apply association rule mining techniques to features associated to the mismatched reviews, to possibly find correlations among review features, scores, and mismatches.

\section*{Acknowledgements}
The research leading to these results has received funding from the European Union under grant 675320 (NECS -- \textit{Network of Excellence in Cybersecurity}) and from the regional project \textit{ReviewLand}, co-funded by Fondazione Cassa di Risparmio di Lucca and IIT-CNR.

\section*{Compliance with Ethical Standards}
\paragraph{\bf Funding} This study was partially funded by the European Union under grant 675320 (NECS -- \textit{Network of Excellence in Cybersecurity}) and by the regional project \textit{ReviewLand}, co-financed by ``Fondazione Cassa di Risparmio di Lucca" and IIT-CNR, (Bando 2016-2017 ``Ricerca'', ReviewLand, IIT-0047044, 29/08/2016).

\paragraph{\bf Conflict of Interest}
Michela Fazzolari, Vittoria Cozza, Marinella Petrocchi and Angelo Spognardi declare that they have no conflict of interest.
\paragraph{\bf Ethical Approval} This article does not contain any studies with the active participation of humans. Furthermore, this article does not contain any studies on animals. The data collected and processed will be solely used for research related to this work and it will be ensured that they will not allow to identify any of the authors of such data.

\end{document}